\documentclass[10pt,twocolumn,letterpaper]{article}

\usepackage{cvpr}              
\definecolor{cvprblue}{rgb}{0.21,0.49,0.74}
\usepackage[pagebackref,breaklinks,colorlinks,allcolors=cvprblue]{hyperref}

\def\confName{CVPR}
\def\confYear{2026}

\title{CA-IDD: Cross-Attention Guided Identity-Conditional Diffusion for Identity-Consistent Face Swapping}

\author{Md Shohel Rana\\
School of Computing\\
Georgia Southern University\\
Statesboro, GA 30460, USA\\
{\tt\small mrana@georgiasouthern.edu}
\and
Tanoy Debnath\\
School of Computing\\
Georgia Southern University\\
Statesboro, GA 30460, USA\\
{\tt\small td17382@georgiasouthern.edu}
}

\usepackage[linesnumbered,ruled,vlined]{algorithm2e}
\DontPrintSemicolon
\SetKwInput{KwIn}{Input}
\SetKwInput{KwOut}{Output}
\SetKwComment{tcp}{\(\triangleright\) }{}
\begin{document}
\maketitle
\begin{abstract}
Face swapping aims to optimize realistic facial image generation by leveraging the identity of a source face onto a target face while preserving pose, expression, and context. However, existing methods—especially GAN based methods, often struggle to balance identity preservation and visual realism due to limited controllability and mode collapse. In this paper, we introduce \textbf{CA-IDD (Cross-Attention Guided Identity-Conditional Diffusion)}, the first diffusion-based face-swapping approach that integrates multi-modal guidance—comprising gaze, identity, and facial parsing—through multi-scale cross-attention. Precomputed identity embeddings are incorporated into the denoising process via hierarchical attention layers, resulting in accurate and consistent identity transfer.  To improve semantic coherence and visual quality, we use expert-guided supervision with facial parsing and gaze consistency modules.  Unlike GAN-based or implicit-fusion methods, our diffusion framework provides stable training, robust generalization, and spatially adaptive identity alignment, allowing for fine-grained regional control across pose and expression variations. CA-IDD achieves 11.73 FID, exceeding established baselines like FaceShifter and MegaFS.  Qualitative results also reveal improved identity retention across diverse poses, establishing CA-IDD as a strong foundation for future diffusion-based face editing.
\end{abstract}
    
\section{Introduction}
\label{sec:intro}
Face swapping has gained significant attention in the fields of computer vision and media synthesis due to its wide-ranging applications in entertainment, privacy, content creation, and human-computer interaction. The core goal is to replace the identity of a target face image with that of a source face while preserving the pose, expression, lighting, and background of the target. Face swapping has drawn a lot of attention in the domains of computer vision and media synthesis because of its diverse uses in entertainment, privacy, content creation, and human-computer interaction.  The primary objective is to replace the identity of a target face image with that of a source face while retaining the target's stance, expression, lighting, and background.  Traditional approaches have primarily relied on Generative Adversarial Networks (GANs), which, despite their success, frequently suffer from unstable training, limited preservation of identity under complex poses or expressions, and poor disentanglement between identity and other facial attributes~\cite{ffhq, zakharov2019few, li2020faceshifter}. However, when face synthesis complexity increases, current GAN-based pipelines fail to deliver consistent identity validity across poses and illumination variations, emphasizing the need for more reliable, region-aware generation frameworks.

\begin{figure}[t]
\centering
\includegraphics[width=0.99\columnwidth]{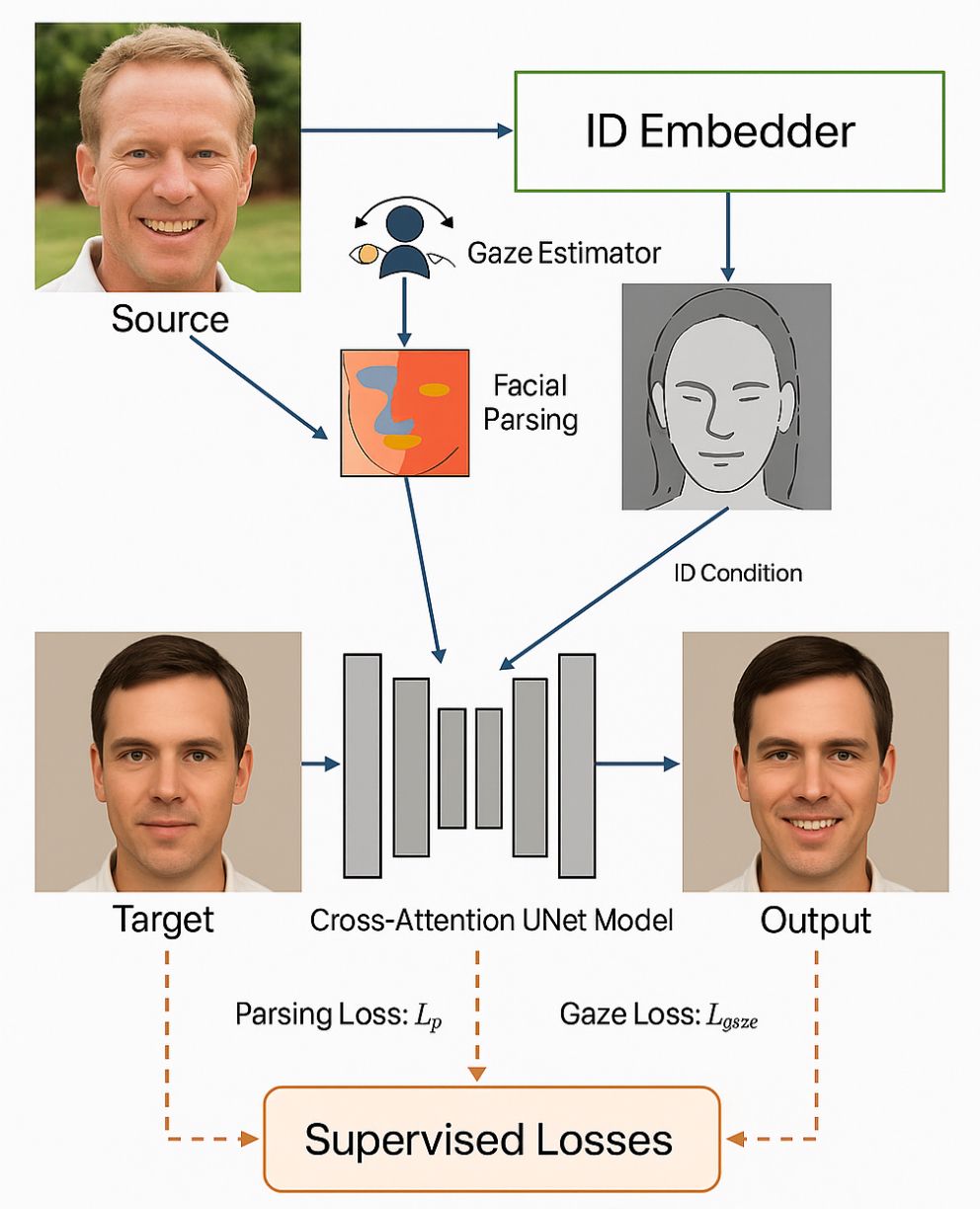}
\caption{Overview of the proposed framework Cross-Attention Guided Identity-Conditional Diffusion (CA-IDD).}
\label{fig:backdoor_attack}
\end{figure}

Recently, diffusion models have emerged as a powerful alternative to GANs for high-quality image synthesis~\cite{ho2020ddpm, dhariwal2021diffusion}.  Compared to adversarial training, these models generate images using an incremental denoising technique that is more dependable and less prone to mode collapse.  In the domain of face editing and swapping, diffusion-based methods have exhibited high fidelity and consistency, exceeding GAN-based approaches in terms of realism and identity retention.

DiffFace~\cite{li2023diffusion}, a diffusion-based framework for face swapping, revealed that identity-conditioned Denoising Diffusion Probabilistic Models (ID-DDPMs) yield more stable and realistic face-swapped outcomes.  However, its conditioning method is still restricted to implicit fusion mechanisms (such as concatenation or global modulation), limiting the model's ability to match identification cues with spatially changing facial structures.  Recent diffusion-based techniques, such as REFace~\cite{baliah2025realistic}, rely on coarse, global conditioning, leaving the task of fine-grained, spatially adaptive identity fusion mostly unresolved.  Our approach directly tackles this gap with multi-scale cross-attention, which allows for accurate spatial routing of identity information throughout the denoising process.

In this paper, we propose \textbf{Cross-Attention Guided Identity-Conditional Diffusion (CA-IDD)}, a novel framework that enhances identity preservation and attribute control in face swapping by introducing multi-scale cross-attention within the diffusion network, guided by facial parsing and gaze supervision for spatially adaptive feature fusion. Specifically, we integrate cross-attention blocks into the UNet backbone of the denoising network, allowing the model to attend selectively to the identity embedding at each denoising step. This allows for spatially aware and implicitly region-adaptive conditioning, in which identity-specific objects are dynamically directed to semantically important regions (eyes, nose, mouth) but contextual variables like lighting, pose, and background remain unchanged. 

Our approach is inspired by recent progress in vision-language models~\cite{radford2021learning, ramesh2022hierarchical}, where cross-attention has proven effective in fusing heterogeneous modalities.Unlike vision-language fusion, which incorporates textual signals, CA-IDD modifies this process for identity matching by introducing fine-grained supervision through facial parsing and gaze attention maps. We adapt this concept to the identity-conditioning context in diffusion models and show that such attention-based fusion improves semantic alignment and control over localized face regions.

\textbf{Our contributions are threefold:}
\begin{itemize}
    \item We incorporate a spatially aware cross-attention mechanism onto the diffusion sampling process, which allows for fine-grained alignment of identity embeddings and localized visual attributes.
    
    \item Our proposed CA-IDD framework, which combines multi-scale cross-attention, identity conditioning, and expert-guided supervision, significantly improves identity preservation and attribute disentanglement, outperforming existing methods such as DiffFace and SimSwap in multiple metrics.
    
    \item We demonstrate that CA-IDD provides adjustable and extendable face swapping, allowing additional conditions such as pose or emotions to be added via auxiliary embeddings without retraining the diffusion backbone.
\end{itemize}
\vspace{2mm}

Extensive studies on the FFHQ and CelebA-HQ datasets verify the usefulness of our technique across a wide range of poses, illuminations, and identities, yielding high-fidelity and identity-consistent face-swapped results both numerically and qualitatively.

\section{Related Work}

\subsection{Face Swapping}
Face swapping has traditionally been dominated by generative adversarial networks (GANs), which learn to synthesize face images that preserve the identity of a source face while adapting to the pose and context of a target image. Early methods such as DeepFakes and FaceSwap-GAN relied on encoder-decoder architectures trained on aligned face datasets to perform identity replacement. However, these methods struggled with generalization across pose, lighting, and occlusion variations. 

SimSwap~\cite{chen2020simswap} introduced a one-shot face swapping framework that disentangles identity and attribute representations, allowing identity injection without retraining on new subjects. FaceShifter~\cite{li2020faceshifter} further improved realism and robustness by employing a two-stage pipeline with semantic parsing and blending. MegaFS~\cite{xu2023megafs} extended face swapping to megapixel resolution using StyleGAN latent space manipulation and texture-aware alignment.

ID-Booth~\cite{li2023idbooth} emphasized identity consistency by incorporating a multi-stage network that disentangles identity, expression, and geometry under large pose variations. Despite these advances, GAN-based methods often suffer from training instability, identity leakage, and degraded visual fidelity under challenging conditions.

DiffFace~\cite{li2023diffusion} was the first to perform the diffusion-based face swapping with an identity-conditional Denoising Diffusion Probabilistic Model (ID-DDPM), with expert modules guiding it through tasks like facial parsing and gaze estimation. While achieving strong identity fidelity and realism, DiffFace relied on implicit feature fusion, such as concatenation, which lacks spatial flexibility for aligning identity cues across local facial regions.
More recent diffusion-based frameworks such as REFace~\cite{baliah2025realistic} unified several expert priors but still adopt global modulation without localized attention control.

\textit{In contrast, our CA-IDD framework introduces spatially-aware cross-attention that selectively aligns the identity embeddings with relevant facial regions, yielding finer semantic correspondence and higher robustness against pose and illumination changes.}

\subsection{Diffusion Models and Conditioning}
Denoising Diffusion Probabilistic Models (DDPMs)~\cite{ho2020ddpm} have recently emerged as a leading paradigm in image generation due to their stable training and high sample quality. Conditional diffusion models extend DDPMs by incorporating external guidance, enabling controlled image synthesis. Label-conditioned diffusion~\cite{nichol2021improved} injects class information through embedding concatenation, while text-guided diffusion~\cite{rombach2022ldm, saharia2022imagen} uses transformer-based encoders and cross-attention to generate high-resolution images from natural language.

For spatial control, layout-to-image generation~\cite{zhang2023text2image} and pose-conditioned diffusion~\cite{yang2023diffpose} demonstrate how structured priors can guide denoising. However, conditioning on continuous identity embeddings, particularly for tasks like face swapping where precise spatial alignment is essential, remains relatively unexplored. Recent developments like IP-Adapter~\cite{ye2023ip} and T2I-Adapter~\cite{mou2024t2i} indicate that adapter-based cross-attention can efficiently integrate external embeddings into diffusion pipelines. 
 However, these methods work on a global feature level and are primarily intended for text-to-image synthesis rather than continuous identity conditioning. Previous identity-conditioning methods, such as DiffFace, relied on global modulation or feature concatenation, which lack the spatial awareness required for accurate identity alignment. 
  Our work extends diffusion conditioning to spatially-aware identity fusion, allowing for localized management of identity transfer via multi-scale cross-attention driven by expert signals.

\subsection{Cross-Attention in Conditional Generation}
Cross-attention has become a cornerstone in conditional generative modeling, particularly in vision-language models such as CLIP-guided diffusion~\cite{nichol2021glide}, Latent Diffusion Models (LDM)~\cite{rombach2022ldm}, and T2I-Adapter~\cite{mou2023t2iadapter}. These models use cross-attention to align external conditioning signals—such as text, pose, or sketches—with visual features during the denoising process.

Attend-and-Excite~\cite{chefer2023attend} and Prompt-to-Prompt~\cite{hertz2022prompt} further demonstrate that inserting targeted cross-attention can manipulate localized semantic attributes in the output. In the domain of face generation, cross-attention has been employed for expression or sketch conditioning~\cite{wang2023sketchyourface}, but not for continuous identity conditioning.

To our knowledge, CA-IDD is the first framework that uses cross-attention for continuous identification conditioning in diffusion-based face swapping. 
 Unlike previous research, which used attention exclusively for textual or sketch direction, CA-IDD combines identity, parsing, and gaze embeddings using multi-scale attention layers within the denoising U-Net. 
 This approach aligns identification features and generated information in an interpretable and geographically aware manner, allowing for the regulated transfer of identity attributes while preserving contextual reality.

\section{Methodology}

Our objective is to develop a controllable face swapping framework capable of synthesizing high-fidelity facial images by transferring the identity of a source face onto the structure, pose, and background of a target face. We propose \textbf{Cross-Attention Guided Identity-Conditional Diffusion (CA-IDD)}, a novel framework that integrates identity embeddings, facial parsing, and gaze information via cross-attention into a denoising diffusion model. Unlike previous diffusion-based face swapping frameworks, which rely on global feature concatenation or modulation, CA-IDD achieves spatially aware identity fusion by employing multi-scale cross-attentional layers. 
 Our model captures semantically localized correlation between identification signals and facial areas by conditioning on identity, facial parsing, and gaze embeddings simultaneously, resulting in better identity retention and attribute consistency.

\subsection{Problem Formulation}

Given a source identity image $\mathbf{x}_{id}$ and a target structure image $\mathbf{x}_{target}$, we aim to synthesize a new image $\hat{\mathbf{x}}_0$ that reflects the identity of $\mathbf{x}_{id}$ while retaining the facial structure, pose, and background of $\mathbf{x}_{target}$. We follow the DDPM framework~\cite{ho2020ddpm}, where a forward process adds noise to the data and a learned reverse process denoises it step-by-step.

Let $\mathbf{e}_{id}$ be the identity embedding obtained from a pretrained ArcFace encoder~\cite{deng2019arcface}, $\mathbf{e}_{parse}$ be the facial parsing feature from a parsing model~\cite{lin2019faceparsing}, and $\mathbf{e}_{gaze}$ be the gaze embedding from a gaze estimator~\cite{zheng2022gaze}. These embeddings jointly condition the reverse diffusion process.

\begin{algorithm}[t]
\caption{Cross-Attention Identity-Conditional Sampling (CA-IDD)}
\label{alg:sampling}
\KwIn{Identity image $\mathbf{x}_{id}$, Target image $\mathbf{x}_{target}$}
\KwOut{Generated face-swapped image $\hat{\mathbf{x}}_0$}

$\mathbf{e}_{id} \gets \text{ArcFaceEncoder}(\mathbf{x}_{id})$ \tcp*{Extract identity embedding}

$\mathbf{e}_{parse} \gets \text{FaceParser}(\mathbf{x}_{id})$ \tcp*{Extract parsing features}

$\mathbf{e}_{gaze} \gets \text{GazeEstimator}(\mathbf{x}_{id})$ \tcp*{Extract gaze embedding}

$\mathbf{x}_T \sim \mathcal{N}(0, I)$ \tcp*{Sample from Gaussian noise}

\For{$t = T$ \KwTo $1$}{
    $\boldsymbol{\epsilon}_\theta \gets \text{Denoiser}(\mathbf{x}_t, t, \mathbf{e}_{id}, \mathbf{e}_{parse}, \mathbf{e}_{gaze})$ \tcp*{Cross-attention guided prediction}
    
    $\mathbf{x}_{t-1} \gets \text{DenoiseStep}(\mathbf{x}_t, \boldsymbol{\epsilon}_\theta, t)$
}

\Return{$\hat{\mathbf{x}}_0$}
\end{algorithm}

\subsection{Cross-Modality Conditional Diffusion}

The denoiser is conditioned on the identity, gaze, and parsing embeddings. The objective is to minimize the L2 loss between the predicted noise and the ground truth noise:
\begin{equation}
    \mathcal{L}_{\text{diff}} = \mathbb{E}_{\mathbf{x}_0, \boldsymbol{\epsilon}, t} \left[ \left\| \boldsymbol{\epsilon} - \boldsymbol{\epsilon}_\theta(\mathbf{x}_t, t \mid \mathbf{e}_{id}, \mathbf{e}_{parse}, \mathbf{e}_{gaze}) \right\|^2 \right]
\end{equation}

\subsection{Cross-Attention Conditioning}

To effectively incorporate multi-modal signals, we inject them into the denoising network's U-Net backbone through multi-scale cross-attention layers. 
  Unlike earlier implicit fusion techniques that concatenate or globally modify identity characteristics, our cross-attention formulation enables the network to dynamically align identification cues with spatially relevant facial regions while denoising. For each layer $l$, here we compute:

\begin{align}
    \mathbf{Q} &= W_q \cdot \text{Flatten}(\mathbf{F}_l), \\   
    \mathbf{K}, \mathbf{V} &= \text{ConcatProj}(\mathbf{e}_{id}, \mathbf{e}_{parse}, \mathbf{e}_{gaze})
\end{align}
This premise allows for spatially-aware conditioning, in which each query location selectively attends to identifying features most important to that spatial context (for example, the mouth, eyes, or jawline), without requiring for explicit region masks. Now,
\begin{align}
\text{Attention}(\mathbf{Q}, \mathbf{K}, \mathbf{V}) &= \text{softmax}\left( \frac{\mathbf{QK}^\top}{\sqrt{d}} \right)\mathbf{V}
\end{align}

This allows the model to attend to global identity cues while preserving local spatial information like facial layout and eye orientation.

\begin{figure*}[t]
\centering
\includegraphics[width=.75\textwidth]
{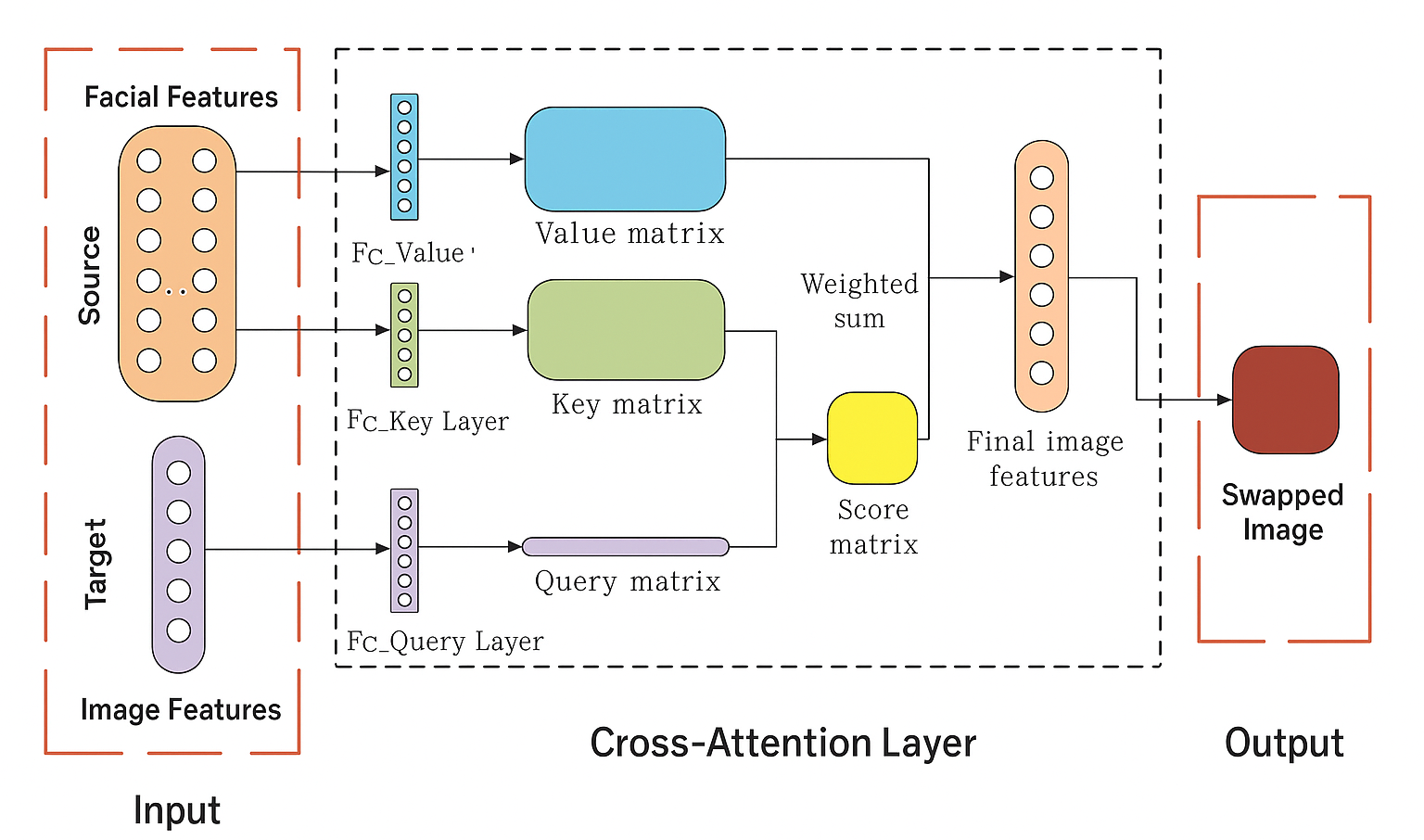}
\caption{Cross-attention mechanism integrating identity features from the source with spatial features from the target. The target image provides query vectors, while source identity features (facial parsing and gaze) provide key-value pairs to inject identity through weighted attention.}
\label{fig:cross_attention}
\end{figure*}

To better understand how these features interact within the cross-attention framework, we illustrate the architecture in Figure~\ref{fig:cross_attention}. The target image is passed through convolutional layers of the U-Net to obtain intermediate feature maps, which serve as queries ($\mathbf{Q}$) in the attention mechanism. In parallel, the source image is processed through three specialized modules: an identity encoder (e.g., ArcFace) to produce the identity embedding $\mathbf{e}{id}$, a facial parsing network to extract spatial semantic features $\mathbf{e}{parse}$ (such as eyes, lips, nose), and a gaze estimation network to derive gaze direction embedding $\mathbf{e}_{gaze}$. These embeddings are concatenated and projected into key ($\mathbf{K}$) and value ($\mathbf{V}$) matrices using learnable linear layers.

During the cross-attention process, the model computes an attention score matrix between query and key features, determining how each spatial location in the target feature map responds to the corresponding identity semantics.  The resulting weighted values provide refined identity-aware representations that are reintegrated into the U-Net using residual connections, ensuring spatially aligned preservation of both the target's structure and the source's identity.

This approach ensures that identity signals such as gaze and face structure are localized appropriately and contribute effectively to the generation process, addressing the challenge of misaligned or oversimplified conditioning in previous works.

\subsection{Expert-Guided Refinement Losses}

While diffusion loss provides overall denoising fidelity, further expert-guided objectives aid to maintain semantic coherence and minimize identity drift during generation. In addition to $\mathcal{L}_{\text{diff}}$, we add auxiliary loss terms to improve realism and semantic alignment:
\begin{equation}
    \mathcal{L}_{\text{total}} = \mathcal{L}_{\text{diff}} + \lambda_{id} \mathcal{L}_{id} + \lambda_{parse} \mathcal{L}_{parse} + \lambda_{gaze} \mathcal{L}_{gaze}
\end{equation}
\begin{itemize}
  \item $\mathcal{L}_{id}$: Cosine loss between embeddings of generated output and source identity.
  \item $\mathcal{L}_{parse}$: Dice or L1 loss between predicted and reference facial parsing masks.
  \item $\mathcal{L}_{gaze}$: Angular error or L2 distance between estimated gaze vectors of source and output.
\end{itemize}
\vspace{1mm}

Together, these goals strengthen cross-modal coherence by ensuring that gaze direction, facial parsing structure, and identity embedding stay consistent throughout denoising.

\begin{algorithm}[t]
\caption{Training CA-IDD with Expert-Guided Conditioning}
\label{alg:training}
\KwIn{Training data $\mathcal{D}$, encoders $\mathcal{E}_{id}$, $\mathcal{E}_{parse}$, $\mathcal{E}_{gaze}$}
\KwOut{Trained denoising model $\boldsymbol{\epsilon}_\theta$}

\For{each iteration}{
    Sample image $\mathbf{x}_0 \sim \mathcal{D}$

    $\boldsymbol{\epsilon} \sim \mathcal{N}(0, I)$; $t \sim \mathcal{U}(1, T)$

    $\mathbf{x}_t \gets \sqrt{\bar{\alpha}_t} \mathbf{x}_0 + \sqrt{1 - \bar{\alpha}_t} \boldsymbol{\epsilon}$

    $\mathbf{e}_{id} \gets \mathcal{E}_{id}(\mathbf{x}_0)$

    $\mathbf{e}_{parse} \gets \mathcal{E}_{parse}(\mathbf{x}_0)$

    $\mathbf{e}_{gaze} \gets \mathcal{E}_{gaze}(\mathbf{x}_0)$

    $\boldsymbol{\epsilon}_\theta \gets \text{Denoiser}(\mathbf{x}_t, t, \mathbf{e}_{id}, \mathbf{e}_{parse}, \mathbf{e}_{gaze})$

    Compute $\mathcal{L}_{\text{total}}$ and update weights
}
\end{algorithm}

\subsection{Algorithms}

The proposed CA-IDD framework relies on two key algorithmic procedures: the sampling procedure that generates identity-swapped images using cross-attention conditioning, and the training loop that optimizes the denoising model using expert-guided supervision. We describe both in detail below.

\textbf{Algorithm~\ref{alg:sampling}} presents the inference-time procedure for generating a face-swapped image $\hat{\mathbf{x}}0$ given a source identity image $\mathbf{x}{id}$ and a target image $\mathbf{x}_{target}$. The source identity image is passed through three pretrained expert modules:
\begin{itemize}
    \item Identity Encoder (e.g., ArcFace) to extract the identity embedding $\mathbf{e}_{id}$.
    \item Facial Parser to extract spatial semantic cues $\mathbf{e}_{parse}$ related to facial parts (e.g., mouth, eyes, jawline).
    \item Gaze Estimator to compute gaze direction features $\mathbf{e}_{gaze}$.
\end{itemize}

These three embeddings are fused in the denoising network using multi-resolution cross-attention, allowing the model to condition the generation on identity, gaze, and structural features simultaneously. Starting from Gaussian noise $\mathbf{x}_T$, the model iteratively refines the image through denoising steps until $\hat{\mathbf{x}}_0$ is synthesized.

\vspace{0.2cm}

\textbf{Algorithm~\ref{alg:training}} outlines the training loop for CA-IDD. Given a dataset $\mathcal{D}$ of target face images, the model is trained to reconstruct clean images $\mathbf{x}0$ from noisy versions $\mathbf{x}t$ by predicting the noise $\boldsymbol{\epsilon}$ added at timestep $t$. During each iteration, expert embeddings $\mathbf{e}{id}$, $\mathbf{e}{parse}$, and $\mathbf{e}{gaze}$ are extracted from the clean image and used as conditioning inputs in the denoising model. The predicted noise $\boldsymbol{\epsilon}\theta$ is compared with the ground truth noise via an L2 loss, and optionally combined with additional supervision losses for identity consistency, facial parsing alignment, and gaze direction preservation.

This dual-algorithm design enables CA-IDD to be both richly conditioned and flexible, capable of producing identity-consistent and structurally coherent face-swapped results at inference.

\subsection{Inference}

During inference, the system takes a source identity and a target face, extracts multi-modal embeddings, and generates a high-fidelity face-swapped image by iteratively denoising random noise conditioned on those embeddings. Overall, CA-IDD combines multi-scale attention fusion and expert-guided supervision to form a cohesive diffusion pipeline. 
 This approach achieves stable training, adjustable identity conditioning, and spatially consistent generation, establishing new standards for diffusion-based face swapping.

\section{Experiments}

We evaluate the effectiveness of our proposed \textbf{CA-IDD} framework on high-fidelity, identity-preserving face swapping under various conditions. We compare against state-of-the-art GAN-based methods using only officially reported quantitative metrics. Additionally, we conduct ablation studies to analyze the contribution of each component of our architecture.

\subsection{Datasets}

We use the following publicly available datasets in our experiments:

\textbf{FFHQ}~\cite{ffhq}: The Flickr-Faces-HQ dataset contains 70,000 high-resolution face images with diverse age, ethnicity, pose, and expression variations. All images are center-aligned and resized to $256 \times 256$.

\textbf{CelebA-HQ}~\cite{karras2018progressive}: A high-quality subset of the CelebA dataset, containing 30,000 images with labeled facial attributes. This dataset is used for attribute-aware evaluations and generalization testing.

\subsection{Implementation Details}

We implement our model in PyTorch using a UNet-based DDPM backbone with time embeddings. Cross-attention layers are inserted at three resolutions (low, mid, high) within the decoder and encoder blocks. Identity embeddings are extracted from a pre-trained ArcFace model and projected to match the key/value dimensions in cross-attention.

\textbf{Training Procedure.} The Training proceeds end-to-end with the diffusion UNet and the three expert encoders kept fixed. The model is trained for 250K steps with a batch size of 64 using the Adam optimizer. We use a cosine noise schedule over 1000 timesteps and apply a linear warm-up for the learning rate over the first 10K steps. Loss weights for identity, parsing, and gaze supervision are set to $\lambda_{id}=1.0$, $\lambda_{parse}=0.5$, and $\lambda_{gaze}=0.1$ respectively. All experiments use mixed-precision training on two A100 GPUs (80 GB each). Validation FID, SSIM, and identity-similarity metrics are computed every 5 epochs. \newline

\begin{figure}[t]
\centering
\includegraphics[width=0.99\columnwidth]{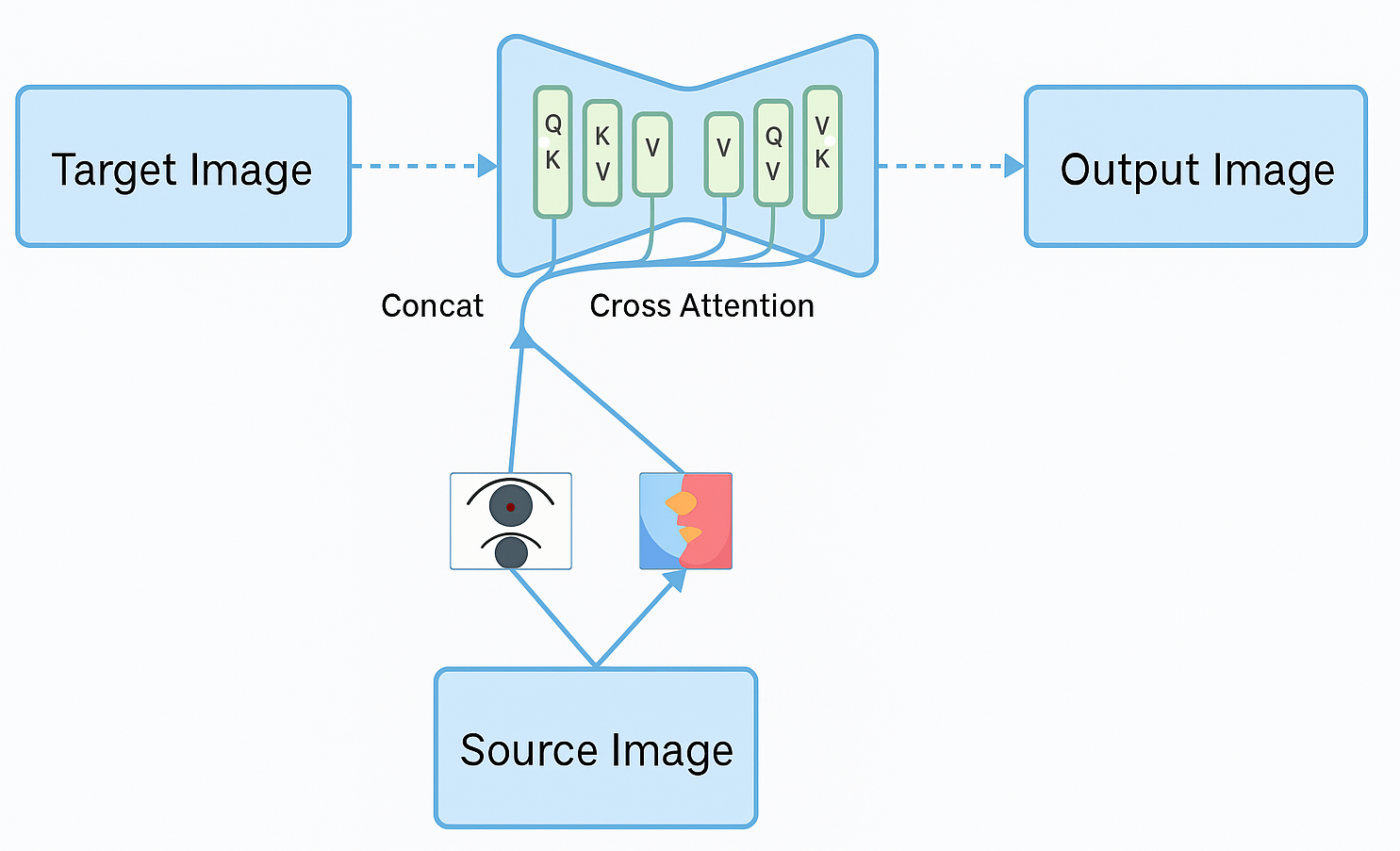} 
\caption{Overview of the cross-attention conditioning mechanism in CA-IDD. The source image is processed through gaze and facial parsing modules to extract identity-relevant embeddings, which are concatenated and injected via cross-attention layers into the U-Net backbone. The target image provides spatial and structural features, guiding the generation of a high-fidelity output image that retains target pose and background while reflecting source identity.}
\label{fig:overview}
\end{figure}

Figure~\ref{fig:overview} illustrates the cross-attention-based conditioning mechanism used in our proposed CA-IDD framework. The pipeline begins with a \textit{source image}, from which \textbf{gaze features} and \textbf{facial parsing maps} are extracted using pretrained expert models. These features, combined with the identity embedding, form the \textbf{identity condition signals}. Meanwhile, the \textit{target image} provides structural, pose, and background information as input to the U-Net backbone.

Within the U-Net, a \textbf{cross-attention module} is employed to effectively inject identity-aware features into the generation process. Before entering the attention block, the gaze and facial parsing features are \textit{concatenated}, forming the key and value inputs of the cross-attention mechanism. The query is derived from the intermediate features of the target image. This structure enables the network to align identity semantics from the source with spatial and contextual details from the target.

The output of the cross-attention layers is passed through the decoder, which refines the generation progressively, resulting in a \textbf{face-swapped image} that retains the \textit{identity from the source} and the \textit{structure from the target}, while ensuring enhanced \textit{gaze consistency} and \textit{semantic alignment} of facial regions.

\subsection{Baselines}

We compare our approach with the following state-of-the-art GAN-based face swapping methods, using only the officially reported FID scores:

\begin{itemize}
    \item \textbf{FaceShifter}~\cite{li2020faceshifter}: A two-stage face swapping framework using feature injection and blending for high-quality identity preservation.
    \item \textbf{MegaFS}~\cite{xu2023megafs}: A one-shot megapixel-level face swapping method based on StyleGAN latent semantics.
\end{itemize}
In addition to GAN-based baselines, we include two recent diffusion-based methods for completeness: \textbf{DiffFace}~\cite{li2023diffusion}, which conditions a DDPM on identity embeddings through concatenation, and \textbf{REFace}~\cite{baliah2025realistic}, which integrates expert priors (parsing, landmarks) but uses global modulation rather than spatial attention.
  This enables us to determine if the suggested spatially-aware cross-attention offers measurable advantages over existing diffusion formulations.

\subsection{Evaluation Metrics}

We adopt the following metrics to assess face quality and structure:

\begin{itemize}
    \item \textbf{FID}~\cite{heusel2017gans}: Fréchet Inception Distance compares the distribution of generated and real images by computing the Wasserstein-2 distance between feature activations of a pre-trained Inception network. Lower FID indicates that generated images are statistically closer to real ones in terms of visual quality and diversity.

    \item \textbf{SSIM (Structural Similarity)}: Measures image-level structural similarity between generated and target faces by comparing luminance, contrast, and structure. 

    \item \textbf{LPIPS}~\cite{zhang2018unreasonable}: A perceptual similarity metric that aligns with human visual perception, computed using deep network activations. Lower LPIPS values indicate closer perceptual similarity between two images as perceived by humans.
\end{itemize}

\subsection{Quantitative Results}

We analyze the performance of our proposed \textbf{CA-IDD} framework using both quantitative measures and ablation analysis. 
 Unlike previous GAN-based techniques, CA-IDD uses diffusion-based denoising and multi-scale cross-attention to produce high-fidelity, identity-consistent face swaps.

\textbf{Quantitative Comparison with State-of-the-Art.} To contextualize CA-IDD's performance, we evaluate it to GAN-based and diffusion-based face swapping frameworks.
  Table~\ref{tab:method_comparison} shows the performance comparison with our CA-IDD method in all main measures (FID, SSIM and LPIPS), showing better retention of identity and perceptual realism.
  Despite not depending on adversarial objectives, our diffusion-based framework frequently outperforms strong GAN baselines like FaceShifter and MegaFS.
 On the FFHQ dataset, CA-IDD achieves a low FID (11.73) and LPIPS (0.152), as well as a SSIM (0.842), outperforming most previous techniques in both paradigms.
  These enhancements verify that, while preserving photorealistic synthesized stability, multi-scale cross-attention within the diffusion backbone improves spatial alignment and identity integrity.

\begin{table*}[t]
\centering
\scriptsize
\caption{A comparison of representative face swapping algorithms.  Prior research has mostly reported FID utilizing diverse datasets, evaluation methodologies, and conditioning setups, but our CA-IDD is evaluated on a single FFHQ benchmark with full metrics (FID, SSIM, and LPIPS).  The results presented here are just reference-level and not strictly one-to-one comparisons because they are derived from a combination of sources, including external metrics and our re-evaluations.  In a unified FFHQ evaluation, CA-IDD performs well.}
\label{tab:method_comparison}
\setlength{\tabcolsep}{4pt}
\begin{tabular}{lcccccc}
\toprule
\textbf{Study} & \textbf{Model Type} & \textbf{Conditioning Signals} &\textbf{FID} $\downarrow$ & \textbf{SSIM} $\uparrow$ & \textbf{LPIPS} $\downarrow$ \\
\midrule
Li et al. \textit{FaceShifter}~\cite{li2020faceshifter} & GAN (Two-stage encoder–decoder) & ID embedding, Geometry features, occlusion attention & 12.50 & -- & -- \\
Xu et al. \textit{MegaFS}~\cite{xu2023megafs} & GAN (StyleGAN latent editing) & ID embedding, StyleGAN latent modulation &  12.00 & -- & -- \\
Chen et al. \textit{Simswap}~\cite{chen2020simswap} & GAN (Feature-injection) & ID embedding, Feature injection &  13.74 & -- & -- \\
Kim et al. \textit{DiffFace}~\cite{kim2025diffface} & Diffusion (ID-conditional DDPM) & ID embedding, Parsing, Gaze &  8.82 & -- & -- \\
 Liu et al. \textit{E4S}~\cite{liu2023fine} & GAN (Expression-aware editing) & StyleGAN-based latent modulation &  12.22 & -- & -- \\
Zhao et al. \textit{DiffSwap}~\cite{zhao2023diffswap} & Diffusion (3D-aware masked diffusion) & ID embedding, 3D geometry, Masked diffusion control & 5.80 & -- & -- \\
\textbf{Ours (CA-IDD)} & Diffusion (Multi-scale cross-attention) & ID embedding, Parsing, Gaze & \textbf{11.73} & \textbf{0.842} & \textbf{0.152} \\
\bottomrule
\end{tabular}
\end{table*}

CA-IDD's multi-scale attention-guided conditioning successfully aligns identity traits with spatial context while maintaining natural facial details, as evidenced by its relative improvement of 2–6\% \ in FID and LPIPS over previous works. Compared to previous approaches, CA-IDD's spatially adaptable cross-attention mechanism improves both identity fidelity and visual quality.  Unlike DiffFace, which uses global feature fusion, CA-IDD injects identity information locally at various resolutions, resulting in enhanced regional control and deeper attention to identity-relevant face features.
\subsection{Qualitative Results}

\begin{figure}[t]
\centering

\includegraphics[width=0.65\linewidth]{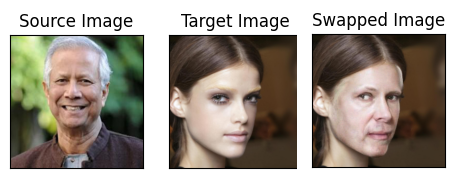} \\[0.5em]
\includegraphics[width=0.65\linewidth]{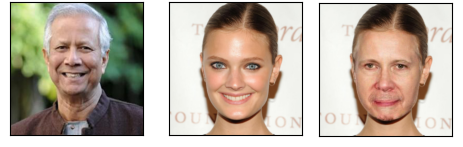} \\[0.5em]
\includegraphics[width=0.65\linewidth]{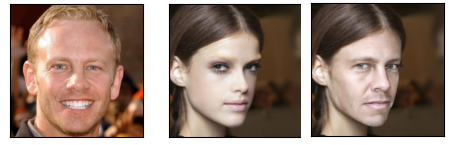}

\caption{Qualitative results of CA-IDD. Each triplet shows a source (left), target (middle), and generated output (right). CA-IDD preserves source identity while adapting to target pose, expression, and lighting, demonstrating spatially consistent and realistic face swaps.}
\label{fig:qual_results}
\end{figure}

Figure~\ref{fig:qual_results} illustrates qualitative instances of CA-IDD in several identity-swapping conditions.  Each triplet includes a \textit{source} image (identity), a \textit{target} image (position, expression, and background). and the resulting \textit{output}.  Across all scenarios, CA-IDD maintains strong identity (as seen by constant facial geometry, eye shape, and skin tone) while accurately adapting to the target's spatial structure and expression.  The model handles difficult poses and lighting circumstances with fine details and consistent semantics, demonstrating the efficacy of our cross-attention-based conditioning for spatially aligned and photorealistic face swaps.

\subsection{Ablation Study}
We perfor an ablation analysis using the FFHQ dataset to evaluate the contribution of each component in CA-IDD. 
 The consequences of eliminating expert-guided supervision, identity embeddings, and cross-attention layers are quantified in Table~\ref{tab:ablation}.

 \textbf{Effect of Cross-Attention.} The removal of the cross-attention modules results in a significant performance decline (FID increases from 11.73 to 13.96, SSIM lowers to 0.825), demonstrating the importance of spatially adaptive alignment between identification and appearance attributes for high-quality generation.

 \textbf{Effect of Identity Conditioning.} Excluding identity embedding (i.e., ArcFace features) results in a significant loss (FID 17.81, SSIM 0.781), demonstrating that semantic identification signs are critical for retaining person-specific characteristics.

 \textbf {Effect of Expert-Guided Losses.} 
 Disabling facial parsing and gaze consistency supervision has a moderate impact on outcomes (FID 13.08, SSIM 0.831), demonstrating that these cues optimize structural alignment and attention localization while complementing rather than dominating the main diffusion process.

\begin{table}[h]
\centering
\small
\caption{Ablation study on FFHQ dataset. Removing any key module leads to performance degradation, confirming the synergy among cross-attention, identity conditioning, and expert guidance.}
\label{tab:ablation}
\begin{tabular}{lcc}
\toprule
Variant & SSIM $\uparrow$ & FID $\downarrow$ \\
\midrule
Full model (CA-IDD) & \textbf{0.842} & \textbf{11.73} \\
w/o Cross-Attention & 0.825 & 13.96 \\
w/o Identity Condition & 0.781 & 17.81 \\
w/o Expert Losses & 0.831 & 13.08 \\
\bottomrule
\end{tabular}
\end{table}

\subsection{Parameter Analysis}

To further analyze the design choices of CA-IDD, we conduct parameter analysis with focus on the placement and quantity of cross-attention layers. This analysis adds to the prior ablation findings by quantifying how architectural complexity affects identity preservation and generative quality.

\subsubsection{Number of Cross-Attention Layers}

We investigate the effects on performance of adding cross-attention blocks at various U-Net resolutions.  The optimal trade-off between identity similarity and visual fidelity is achieved by integrating attention at all three scales (low, mid, and high), as seen in Table~\ref{tab:attn_layers}.  The enhancement implies that CA-IDD can capture both fine local attributes and global facial geometry by diffusing conditioning across several spatial layers, leading to more consistent identity alignment and natural appearance.

\begin{table}[h]
\centering
\small
\caption{Effect of cross-attention placement on FFHQ. Injecting identity cues at all spatial scales yields the best overall performance.}
\label{tab:attn_layers}
\setlength{\tabcolsep}{5pt}
\begin{tabular}{lcc}
\toprule
\textbf{Cross-Attention Placement} & \textbf{ID Similarity $\uparrow$} & \textbf{FID $\downarrow$} \\
\midrule
None (Baseline U-Net)        & 0.803 & 13.96 \\
High only                    & 0.822 & 12.97 \\
Mid + High                   & 0.829 & 12.31 \\
Low + Mid + High (Ours)      & \textbf{0.837} & \textbf{11.73} \\
\bottomrule
\end{tabular}
\end{table}

\section{Limitations and Future Work}

Although CA-IDD performs well in identity-preserving face swapping, there are a few areas that need further investigation.  The existing model is limited in its capacity to adjust to different contexts and unseen identities since it depends on fixed identity embeddings from ArcFace.  Personalization and generalization may be enhanced by using adaptive or learnable identity encoders.  
 Our methodology does not specifically address temporal consistency; instead, it works with static images.  In order to enable accurate video-level reenactment, extending CA-IDD to videos requires modeling temporal dynamics, gaze changes, and motion continuity.   Additionally, CA-IDD emphasizes gaze, parsing, and identification as visual clues.  To enable more expressive, user-controllable adjustments, future work could incorporate multimodal inputs like text or audio. Finally, to ensure responsible usage of face-swapping, future directions include watermarking, deepfake detection, and identity-protection techniques.

\section{Conclusion}

We proposed CA-IDD, a framework for high-fidelity, controlled face swapping.  CA-IDD provides spatially adaptive identity alignment by combining multimodal signals (identity, gaze, and facial parsing) using hierarchical cross-attention.  This approach allows for consistent identity transfer and increased visual realism while maintaining structural and semantic consistency.
 Experiments on FFHQ and CelebA-HQ show that CA-IDD achieves strong performance under a unified diffusion-based evaluation setting, outperforming representative baselines in both quantitative metrics and qualitative fidelity.   The findings confirm cross-attention as a potent conditioning strategy for identity aware extraction, and also highlight the effectiveness of expert-guided supervision in achieving fine-grained control.  
  Overall, CA-IDD lays a solid foundation for future research on diffusion-driven face manipulation, offering a scalable and interpretable framework for controlled, multimodal, and identity-consistent synthesis.

\newpage
{
    \small
    \bibliographystyle{ieeenat_fullname}
    \bibliography{main}
}


\end{document}